\documentclass{article}
\usepackage{microtype}
\usepackage{graphicx}
\usepackage{booktabs}
\usepackage{multirow}
\usepackage{amsmath,amssymb,amsfonts,mathtools}
\usepackage{xcolor}
\usepackage{wasysym}
\usepackage{hyperref}
\usepackage{float}

\usepackage[acronym]{glossaries}

\usepackage[accepted]{mlsys2025_arxiv} 
\makeatletter
\renewcommand{\Notice@String}{Preprint. Under review at the Conference on Machine Learning and Systems (MLSys 2027)}
\makeatother

\usepackage{enumitem}

\setlist{
    nosep,
    topsep=2pt,
    partopsep=0pt,
    itemsep=1pt,
    parsep=0pt
}

\newcommand{\U}{\mathcal{U}}
\newcommand{\T}{\mathcal{T}}
\newcommand{\Sc}{\mathcal{S}}
\newcommand{\G}{\mathcal{G}}
\newcommand{\Rb}{\mathcal{R}}
\newcommand{\K}{\mathcal{K}}
\newcommand{\ind}{\mathbf{1}}
\newcommand{\thout}{\theta^{\mathrm{out}}}
\newcommand{\thin}{\theta^{\mathrm{in}}}
\newcommand{\Scale}{\textsc{Scale}}
\newcommand{\Cond}{\textsc{Conditional}}
\newcommand{\Reject}{\textsc{Reject}}
\newcommand{\evid}[1]{{\footnotesize\textcolor{gray}{[#1]}}}

\mlsystitlerunning{EnterpriseVal: Quantifying the Efficacy, Reliability and Value of Generative AI in the Enterprise}

\makeglossaries

\begin{document}

% abbreviations
\newacronym{Al-Jazeera}{Al-Jazeera}{Al-Jazeera Satellite Channel}
\newacronym{AraGPT-2}{AraGPT-2}{Arabic Generative Pre-trained Transformer 2}
\newacronym{AraGPT-3}{AraGPT-3}{Arabic Generative Pre-trained Transformer 3}
\newacronym{ArSAS}{ArSAS}{Arabic Speech-Act and Sentiment}
\newacronym{PATB}{PATB}{Penn Arabic Treebank}
\newacronym{BBC}{BBC}{British Broadcasting Corporation}
\newacronym{BERT}{BERT}{Bidirectional Encoder Representations from Transformers}
\newacronym{BPE}{BPE}{byte-pair encoding}
\newacronym{mBERT}{mBERT}{multi-lingual BERT}
\newacronym{RoBERTa}{RoBERTa}{A Robustly Optimized BERT Pre-training Approach}
\newacronym{CA}{CA}{Classical Arabic}
\newacronym{CC}{CC}{Common Crawl}
\newacronym{CC-100}{CC-100}{Common Crawl 100}
\newacronym{CNN}{CNN}{Cable News Network}
\newacronym{DA}{DA}{Dialectal Arabic}
\newacronym{EAPCOUNT}{EAPCOUNT}{English-Arabic Parallel Corpus of the United Nations Texts}
\newacronym{GenAI}{GenAI}{Generative AI}
\newacronym{eVal}{eVal}{EnterpriseVal}
\newacronym{GPT-1}{GPT-1}{Generative Pre-trained Transformer 1}
\newacronym{GPT-2}{GPT-2}{Generative Pre-trained Transformer 2}
\newacronym{GPT-3}{GPT-3}{Generative Pre-trained Transformer 3}
\newacronym{GPT}{GPT}{Generative Pre-trained Transformer}
\newacronym{GPU}{GPU}{Graphics Processing Unit}
\newacronym{HPL}{HPL}{High-Performance Linpack} 
\newacronym{IR}{IR}{Information Retrieval} 
\newacronym{KSUCCA}{KSUCCA}{King Saud University Corpus of Classical Arabic}
\newacronym{LM}{LM}{Language model}
\newacronym{LLM}{LLM}{Large Language Model}
\newacronym{LSTM}{LSTM}{Long-Short Term Memory}
\newacronym{ME}{ME}{Middle East}
\newacronym{MGB-2}{MGB-2}{Multi-Genre Broadcast-2}
\newacronym{MSA}{MSA}{Modern Standard Arabic}
\newacronym{MEGLN}{MEGLN}{MSA, Egyptian, Gulf, Levant \& North African}
\newacronym{MRC}{MRC}{Machine Reading Comprehension}
\newacronym{MT}{MT}{Machine Translation}
\newacronym{NN}{NN}{Neural Network}
\newacronym{NQ}{NQ}{Natural Questions}
\newacronym{NLP}{NLP}{Natural Language Processing}
\newacronym{OpenITI}{OpenITI}{Open Islamicate Texts Initiative}
\newacronym{OSAC}{OSAC}{Open-Source Arabic Corpora}
\newacronym{OSCAR}{OSCAR}{Open Super-large Crawled ALMAnaCH coRpus}
\newacronym{OSIAN}{OSIAN}{Open Source International Arabic News}
\newacronym{POS}{POS}{Part-of-speech}
\newacronym{PTB}{PTB}{Penn Tree Bank}
\newacronym{QA}{QA}{Question Answering} 
\newacronym{RNN}{RNN}{Recurrent Neural Network}
\newacronym{RT}{RT}{Russia Today}
\newacronym{SM}{SM}{Sentence Match}
\newacronym{SQuAD}{SQuAD}{Stanford Question Answering Dataset}
\newacronym{TPU}{TPU}{Tensor Processing Unit}
\newacronym{TFRC}{TFRC}{TensorFlow Research Cloud}
\newacronym{KSU}{KSU}{King Saud University}
\newacronym{WER}{WER}{Word Error Rate}
\newacronym{XNLI}{XNLI}{Cross-lingual Natural Language Inference}
\newacronym{HARD}{HARD}{Hotel Arabic Reviews Dataset}

\twocolumn[
\mlsystitle{EnterpriseVal: Quantifying the Efficacy, Reliability and Value of Generative AI in the Enterprise}

\mlsyssetsymbol{equal}{*}
\begin{mlsysauthorlist}
\mlsysauthor{Abbas Raza Ali}{equal,abbas}
\mlsysauthor{Muhammad Ajmal Siddiqui}{ajmal}
\mlsysauthor{Moona Zahid}{moona}
\end{mlsysauthorlist}

\mlsysaffiliation{abbas}{Citigroup Inc., London, United Kingdom}
\mlsysaffiliation{ajmal}{Ernst \& Young LLP, London, United Kingdom}
\mlsysaffiliation{moona}{NVIDIA Corporation, London, United Kingdom}

\mlsyscorrespondingauthor{Abbas Raza Ali}{abbas.raza.ali@gmail.com}
\mlsyskeywords{Evaluation, Generative AI, Productivity, Enterprise deployment}

\vskip 0.3in

\begin{abstract}
Frontier language models now produce professional deliverables that expert graders judge to match human work on a substantial share of economically valuable tasks, yet most enterprise GenAI initiatives fail to show a measurable business effect and a large fraction of agentic projects are expected to be cancelled. We argue that this is substantially a \emph{measurement} problem: public benchmarks answer ``what can the model do?'', whereas a deployment decision requires ``is this workflow fit, reliable, safe and worth scaling - here, on our data, under our controls?''. We present \textbf{EnterpriseVal}, a use-case-level evaluation system that closes this gap. It comprises (i)~a formal specification of the use case and of the frozen socio-technical configuration under test, model, prompts, retrieval, tools, guardrails and human oversight, with an autonomy level and consequence tier that jointly set the required evaluation intensity; (ii)~a metric catalogue spanning fidelity, utility, efficiency, reliability, assurance and oversight; (iii)~a grading protocol that scales blinded expert judgement with calibrated LLM-as-judge scoring through prediction-powered inference; (iv)~a two-tier threshold gate, stated as an executable algorithm, that maps metric vectors with confidence bounds to \Reject/\Cond/\Scale{} decisions; and (v)~a value-and-risk model in which the reviewer catch rate is a measured parameter. We report a pilot across three workflows in a global bank. In credit-memo drafting, human-graded citation precision reached 88\% and hallucination rate 1.6\% for the best model against gates of 70\% and 5\%; in procedure transformation, analyst refinement effort fell from an estimated 27.4 to 2.9 hours per document. We separate established results, documented pilot evidence, the proposed system and open hypotheses, and specify the experiments required for full validation.
\end{abstract}
]

\printAffiliationsAndNotice{}

%%%%%%%%%%%%%%%%%%%%%%%%%%%%%%%%%%%%%%%%%%%%%%%%%%%%%%%%%%%%%%%%%%%%%%%%%%%%%%%%
\section{Introduction}
\label{sec:intro}

The central challenge for enterprise AI is no longer whether foundation models can perform valuable professional work, but whether that capability can be translated into reliable, measurable, and economically sustainable enterprise outcomes. Three developments make this question increasingly urgent. First, model capability on realistic professional work has risen steeply: on GDPval, 1{,}320 expert-authored tasks across 44 occupations, the best frontier models in late 2025 produced deliverables rated as good as or better than experienced professionals in just under half of blinded pairwise comparisons, and performance more than tripled between GPT-4o and GPT-5~\cite{gdpval}. Second, realised enterprise value lags far behind: an MIT study of roughly 300 deployments found about 95\% of pilots with no measurable P\&L impact~\cite{mit_divide}, and McKinsey's 2025 survey found 88\% of organisations using AI in at least one function but only 39\% attributing any EBIT impact to it~\cite{mckinsey2025}. Third, agentic systems are arriving before the evaluation discipline needed to operate them: Gartner projects that over 40\% of agentic AI projects will be cancelled by 2027 for unclear value or inadequate risk controls~\cite{gartner2025}, and a survey of 306 practitioners running agents in production finds reliability to be the dominant barrier, with most restricting agents to a few steps before human intervention~\cite{pan2025agents}.

\begin{figure}[t]
\centering
\includegraphics[width=1.07\columnwidth]{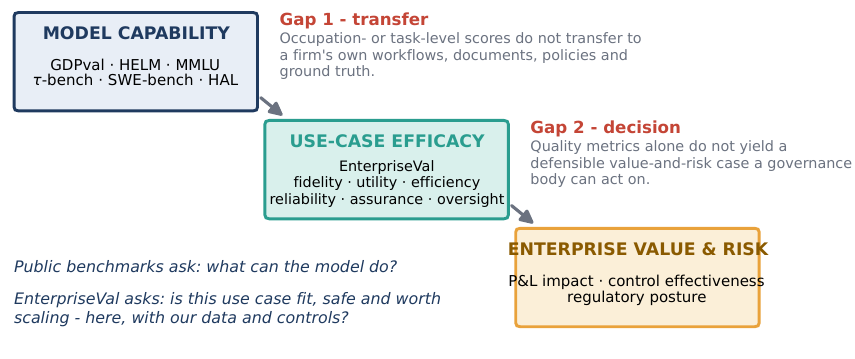}
\caption{\textbf{Two gaps between public benchmarks and enterprise decisions.} Capability benchmarks measure what a model can do on portable tasks. Enterprises must also establish transfer to their own workflows and ground truth (Gap~1) and convert quality metrics into a value-and-risk case a governance body can act on (Gap~2).}
\label{fig:gap}
\end{figure}

We contend that much of the distance between the first fact and the second is a measurement gap with two parts (Figure~\ref{fig:gap}). Capability benchmarks are built to be portable across organisations, and that portability is exactly what prevents them from answering the questions that decide a deployment: whether a credit analyst may rely on a generated repayment analysis depends on the firm's documents, credit policy, citation conventions and reviewer capacity. Conversely, the metrics that internal teams do compute rarely reach a governance committee in a decision-grade form: thresholds are set after the fact, sample sizes go unreported, LLM-graded scores are presented without calibration, and the link from metric to money is asserted rather than modelled.

\gls{eVal} treats evaluation as a systems problem: a repeatable pipeline that takes a frozen workflow configuration and a firm-specific task set and emits an auditable gate decision, an efficacy index and a value estimate with stated uncertainty. It sits between model benchmarks such as GDPval~\cite{gdpval}, HELM~\cite{helm} and the agent suites $\tau$-bench~\cite{taubench,tau2bench}, SWE-bench~\cite{swebench} and HAL~\cite{hal}, and the model-risk-management regimes of regulated industries~\cite{ss123}. It borrows GDPval's blinded expert comparison, the reliability and cost metrics of the agent-evaluation literature~\cite{taubench,kapoor2024agents,rabanser2026reliability}, and the statistical hygiene of survey methodology, and adds decision rules and a value model that a risk committee can act on.

\paragraph{Contributions.}
\begin{enumerate}\setlength\itemsep{0pt}
\item A formal evaluation object, use case $\U$ and frozen configuration $\sigma$ including human oversight, with an \emph{autonomy}$\times$\emph{consequence} evaluation-intensity matrix (\S\ref{sec:framework}).
\item A six-family metric catalogue with operational definitions, extending pilot fidelity metrics with reliability (pass\textasciicircum{}$k$), assurance and oversight metrics (\S\ref{sec:metrics}).
\item A grading and statistical protocol: blinded experts with agreement targets, calibrated LLM-as-judge with prediction-powered correction, sample-size rules, configuration freeze and harness parity (\S\ref{sec:grading}).
\item A two-tier gate and Efficacy Index, given as algorithms, and a value-and-risk model with the reviewer catch rate as an explicit parameter (\S\ref{sec:decision}).
\item Documented pilot evidence from three workflows in a global bank, reported with limitations, and the experiments needed to complete validation (\S\ref{sec:pilot}--\S\ref{sec:limits}).
\end{enumerate}
Throughout we label claims as \evid{Established} (cited prior work), \evid{Documented} (our pilot), \evid{Proposed} (our design) or \evid{Hypothesis}.

%%%%%%%%%%%%%%%%%%%%%%%%%%%%%%%%%%%%%%%%%%%%%%%%%%%%%%%%%%%%%%%%%%%%%%%%%%%%%%%%
\section{Background and Related Work}
\label{sec:related}
\evid{Established}

\paragraph{From capability to economically valuable tasks.} Knowledge tests such as MMLU~\cite{mmlu} saturated quickly and suffer from contamination and weak construct validity~\cite{eriksson2025trust,raji2021}. HELM broadened coverage but stayed model-centric~\cite{helm}. GDPval shifted to expert-authored, expert-graded professional deliverables and reported that reasoning effort and task context each yield measurable gains and that models complete tasks roughly $100\times$ faster and cheaper than experts~\cite{gdpval}. GDPval is organised by occupation and designed to be portable; \gls{eVal} is organised by use case and designed to be firm-specific. They are complementary: GDPval shortlists model families, \gls{eVal} decides deployments.

\paragraph{Agent evaluation and the reliability gap.} $\tau$-bench introduced pass\textasciicircum{}$k$, success on all $k$ attempts, and showed agents that succeed 60\% of the time once may succeed only 25\% of the time on all of eight tries~\cite{taubench}. Across 24 months of frontier releases, reliability improved far less than accuracy and plateaued similarly across vendors~\cite{rabanser2026reliability}. Accuracy without cost yields needlessly expensive agents~\cite{kapoor2024agents}; HAL standardises cost logging~\cite{hal}, and the CLEAR framework adds latency, assurance and reliability, finding that expert judgement of production success tracks a multidimensional score far better than accuracy~\cite{mehta2025clear}. KAMI's 5.5-billion-token study found public leaderboard rank a poor predictor of performance on routine enterprise tasks, with small prompt and tool-message changes moving accuracy and cost by large factors~\cite{roig2025kami}; log-level analysis is increasingly regarded as necessary~\cite{loganalysis2026}. Enterprise sandboxes~\cite{vishwakarma2025sandbox,wang2025enterprise} remain generic across firms. A systematic review of 84 agentic-AI evaluations finds capability measured extensively while workflow integration, trust and operation over time are rarely measured~\cite{measurement_imbalance}, and predictive validity, whether in-sample rankings transfer, has been proposed as the standard for agent leaderboards~\cite{predictive_validity}.

\paragraph{Grounded generation.} Enterprise drafting is retrieval-augmented generation~\cite{rag}, and its principal failure is unfaithful content~\cite{huang2025hallucination,ji2023hallucination}. FActScore checks atomic facts against a source~\cite{factscore}; RAGAS provides reference-free faithfulness and relevance~\cite{ragas}. LLM-as-judge methods correlate well with human preference~\cite{mtbench,geval} but exhibit position, verbosity and self-preference biases~\cite{gu2024judge}, and aligning judges with human criteria is itself iterative~\cite{shankar2024validators}.

\paragraph{Productivity and oversight.} Field studies report 14\% more issues resolved per hour in customer support~\cite{Abbas2024}, 40\% faster professional writing~\cite{noy2023}, and 12\% more tasks for consultants alongside \emph{lower} accuracy outside model competence - the ``jagged frontier''~\cite{dellacqua2023}. Time saved must therefore be measured net of review, and review performance itself must be measured.

\paragraph{Governance.} In banking, quantitative methods fall under model risk management (SR~11-7, SS1/23)~\cite{ss123}; the NIST AI RMF and its GenAI Profile call for measurement, thresholds and documentation~\cite{nistgenai}; the EU AI Act imposes risk-management, data-governance, human-oversight, accuracy and logging duties on high-risk systems~\cite{euaiact}, with stand-alone high-risk obligations deferred to December 2027 by the 2026 Digital Omnibus~\cite{euomnibus}; ISO/IEC 42001 and 42005 provide management-system and impact-assessment standards~\cite{iso42001,iso42005}. No prior framework combines firm-specific ground truth, decision thresholds with uncertainty, a value model and a governance mapping (Table~\ref{tab:positioning}).

\begin{table}[t]
\centering\footnotesize
\caption{Positioning against representative frameworks (\CIRCLE\ full, \LEFTcircle\ partial, \Circle\ absent).}
\label{tab:positioning}
\setlength{\tabcolsep}{3pt}
\resizebox{\columnwidth}{!}{%
\begin{tabular}{@{}lcccccc@{}}
\toprule
Dimension & GDPval & $\tau$/HAL & CLEAR & KAMI & RAGAS & \textbf{EV} \\
\midrule
Firm-specific ground truth & \Circle & \Circle & \Circle & \Circle & \LEFTcircle & \CIRCLE \\
Blinded expert grading & \CIRCLE & \Circle & \LEFTcircle & \Circle & \Circle & \CIRCLE \\
Calibrated LLM judge & \LEFTcircle & \Circle & \LEFTcircle & \Circle & \CIRCLE & \CIRCLE \\
Reliability (pass\textasciicircum{}$k$) & \Circle & \CIRCLE & \CIRCLE & \CIRCLE & \Circle & \CIRCLE \\
Cost / efficiency & \CIRCLE & \CIRCLE & \CIRCLE & \CIRCLE & \Circle & \CIRCLE \\
Assurance / oversight & \Circle & \LEFTcircle & \CIRCLE & \Circle & \Circle & \CIRCLE \\
Thresholds with CIs & \Circle & \Circle & \Circle & \LEFTcircle & \Circle & \CIRCLE \\
Value-and-risk model & \Circle & \Circle & \LEFTcircle & \Circle & \Circle & \CIRCLE \\
Governance mapping & \Circle & \Circle & \Circle & \Circle & \Circle & \CIRCLE \\
\bottomrule
\end{tabular}}
\end{table}

%%%%%%%%%%%%%%%%%%%%%%%%%%%%%%%%%%%%%%%%%%%%%%%%%%%%%%%%%%%%%%%%%%%%%%%%%%%%%%%%
\section{The EnterpriseVal Framework}
\label{sec:framework}
\evid{Proposed}

\subsection{Design principles}
Five principles, each traceable to a failure in \S\ref{sec:related}: \emph{evaluate the workflow, not the model}; \emph{ground truth is expert-authored and firm-specific}; \emph{every score carries its uncertainty and its grader}; \emph{thresholds are decisions set before results are seen}; and \emph{value is modelled, not asserted}. A sixth is a systems requirement: \emph{the configuration is frozen and human oversight is measured} - comparisons are valid only under harness parity, and ``human in the loop'' is a component whose detection performance is measured, not an assumption that converts an imperfect system into a safe one.

\subsection{Formal setting}
A use case is a tuple
\begin{equation}
\U=\langle \T,\Sc,\G,\Rb,a,c\rangle,\qquad t_i=(x_i,S_i,g_i)\in\T,
\label{eq:usecase}
\end{equation}
where $\T$ is a set of $n$ tasks sampled from the workflow population, each an instruction $x_i$, permitted sources $S_i$ and an expert gold artefact $g_i$; $\Sc$ the corpus and access policy; $\G$ the gold-set specification including the \emph{key elements} $\K(g_i)$ a competent artefact must contain; $\Rb$ the rubric; $a\in\{\mathrm{A0},\dots,\mathrm{A4}\}$ the autonomy level and $c\in\{\mathrm{C1},\dots,\mathrm{C4}\}$ the consequence tier (\S\ref{sec:autonomy}). The candidate system is itself a tuple, frozen for the run:
\begin{equation}
\sigma=\langle M,P,\Phi,U,\Gamma,H\rangle,\qquad (y_i,\tau_i)=\sigma(x_i,S_i),
\label{eq:config}
\end{equation}
with $M$ the model(s) and sampling settings, $P$ prompts and policies, $\Phi$ retrieval, memory and context management, $U$ tools and permissions, $\Gamma$ guardrails, verification and logging, and $H$ human roles, checkpoints and escalation rules; $\tau_i$ is the trajectory of tool calls and intermediate states. Making $H$ part of $\sigma$ is deliberate: two deployments of one model with different review designs are different systems. Any change to $\sigma$ is a re-evaluation trigger. The evaluation returns a metric vector $\mathbf m(\sigma,\U)$ with confidence intervals and a gate decision.

\subsection{Architecture}
\begin{figure}[t]
\centering
\includegraphics[width=1.03\columnwidth]{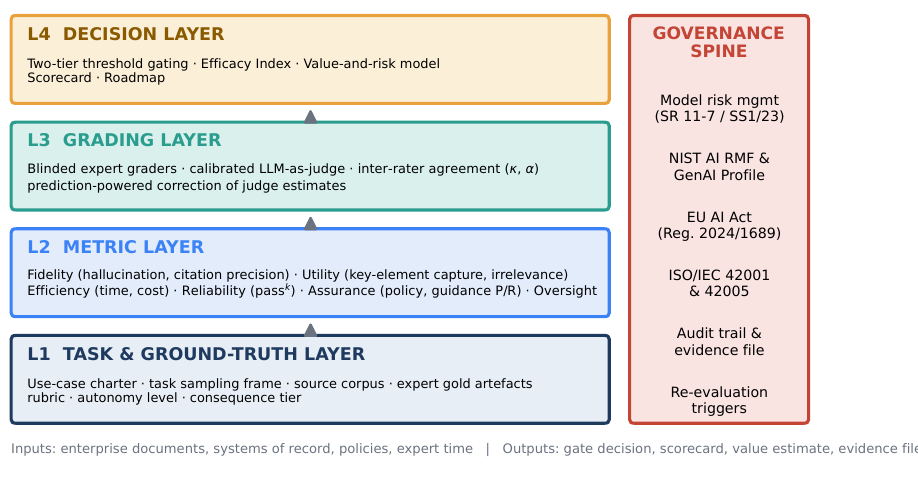}
\caption{\textbf{\gls{eVal} architecture.} Four evaluation layers feed a governance decision; the governance spine consumes artefacts from every layer.}
\label{fig:arch}
\end{figure}

Figure~\ref{fig:arch} shows four layers and a governance spine. Layer~1 constructs $\U$; Layer~2 computes metrics (\S\ref{sec:metrics}); Layer~3 produces graded judgements with uncertainty (\S\ref{sec:grading}); Layer~4 applies thresholds and the value model (\S\ref{sec:decision}). The spine maps each artefact to MRM, NIST, EU AI Act and ISO~42001 requirements so the evaluation doubles as the model-risk evidence file (Appendix~C (supplementary)).

\subsection{Autonomy, consequence and evaluation intensity}
\label{sec:autonomy}
\begin{figure}[t]
\centering
\includegraphics[width=1.03\columnwidth]{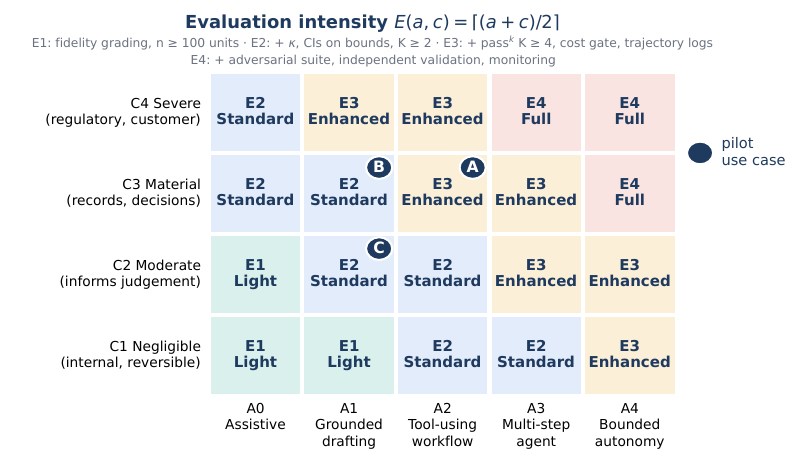}
\caption{\textbf{Evaluation-intensity matrix} (Eq.~\ref{eq:intensity}). Markers place the pilot use cases: B credit memos (A1,C3), A control assessment (A2,C3), C procedure transformation (A1,C2).}
\label{fig:matrix}
\end{figure}

Autonomy describes what a system \emph{does}; consequence describes what is at stake when it is wrong. We define five autonomy levels, adapted from~\cite{morris2024levels}: A0 assistive Q\&A; A1 grounded drafting with citations and human edits; A2 tool-using workflow (retrieval, extraction, rule checks); A3 multi-step agent with human checkpoints; A4 bounded autonomy acting on systems with post-hoc review. Each level inherits all lower-level metrics and adds families (Table~\ref{tab:metrics}). Orthogonally, consequence tiers run from C1 (internal, reversible) through C2 (informs judgement) and C3 (alters records or decisions) to C4 (regulatory, customer or financial harm), following MRM materiality logic~\cite{ss123} and ISO/IEC~42005~\cite{iso42005}. The two set the \emph{evaluation intensity}
\begin{equation}
E(a,c)=\left\lceil \tfrac{a+c}{2}\right\rceil\in\{1,\dots,4\},
\label{eq:intensity}
\end{equation}
with $a\in\{0..4\}$, $c\in\{1..4\}$ (Figure~\ref{fig:matrix}). E1 requires fidelity grading on $\ge$100 units; E2 adds inter-rater agreement, CIs with gate decisions on conservative bounds and $K\ge2$ repeats; E3 adds pass\textasciicircum{}$k$ with $K\ge4$, a cost-per-successful-task gate and trajectory logging; E4 adds an adversarial suite, independent validation and production monitoring. A low-autonomy, high-consequence drafting assistant is thus not under-evaluated because it ``only drafts''.

%%%%%%%%%%%%%%%%%%%%%%%%%%%%%%%%%%%%%%%%%%%%%%%%%%%%%%%%%%%%%%%%%%%%%%%%%%%%%%%%
\section{Metric Layer}
\label{sec:metrics}
\evid{Proposed; individual metrics instantiate established constructs~\cite{factscore,ragas,taubench}}

Metrics are grouped into six families answering distinct decision questions - \emph{can I trust it} (fidelity), \emph{is it useful} (utility), \emph{does it save effort} (efficiency), \emph{will it keep working} (reliability), \emph{does it stay within the rules} (assurance) and \emph{does the human catch what it misses} (oversight). Units are those a reviewer inspects: sentences, cited chunks, key elements, tasks, trajectories (Table~\ref{tab:metrics}).

\paragraph{Fidelity.} For output $y$ segmented into sentences, the hallucination rate is the share of assertive sentences unsupported by the permitted sources,
\begin{equation}
\mathrm{HR}(y)=\tfrac{1}{|y|}\textstyle\sum_{s\in y}\ind[\,\mathrm{assert}(s)\wedge\neg\,\mathrm{supp}(s\mid S)\,],
\label{eq:hr}
\end{equation}
a \emph{faithfulness} rather than world-factuality definition~\cite{huang2025hallucination}: a true statement from a non-permitted source counts, because the workflow requires traceability. The sentence is the operational unit; the reference definition is at the level of atomic claims~\cite{factscore}, and the two are reconciled on a calibration subset because one sentence may carry several claims. Citation precision is the share of cited chunks actually used by the citing paragraph, $\mathrm{CP}=|\{c:\mathrm{used}(c,\mathrm{para}(c))\}|/|\mathcal C(y)|$; the pilot used a \emph{lenient} criterion (any part reflected), and we recommend also reporting a strict variant and citation recall.

\paragraph{Utility.} Key-element capture $\mathrm{KEC}=|\K(g)\cap\K(y)|/|\K(g)|$ is a recall measure against expert expectation; irrelevance rate $\mathrm{IR}=\tfrac1{|y|}\sum_s \ind[s\notin\mathrm{rel}(x)]$ is a precision-like measure; reporting both guards against a system that captures everything by writing everything.

\paragraph{Efficiency.} Net of review: $\Delta_{\mathrm{eff}}=1-\bar t_{\mathrm{AI}}/\bar t_{\mathrm{manual}}$ with $\bar t_{\mathrm{AI}}=\bar t_{\mathrm{gen}}+\bar t_{\mathrm{review}}$. \emph{Estimated} baselines (practitioner recall) are adequate for prioritisation; \emph{observed} baselines from timed matched tasks~\cite{noy2023,dellacqua2023} are required for a value claim. Cost per \emph{successful} task is recorded alongside~\cite{kapoor2024agents}.

\paragraph{Reliability.} With $K$ attempts per task and $c_i$ successes,
\begin{equation}
\widehat{\mathrm{pass}^k}=\tfrac1n\textstyle\sum_{i=1}^{n}\binom{c_i}{k}\big/\binom{K}{k},
\label{eq:passk}
\end{equation}
the unbiased estimator of $\Pr[\text{all }k\text{ attempts succeed}]$~\cite{taubench}. We report pass\textasciicircum{}1, 2 and 4; the decay is the reliability signature that has stagnated across frontier models~\cite{rabanser2026reliability}.

\paragraph{Assurance.} Precision and recall of a \emph{guidance verification} step against mandatory policy items (recall weighted more, since a missed requirement costs more than a false alarm); for A3+ systems, policy-violation rate, escalation appropriateness and incidents under an adversarial suite of injected instructions in tool outputs~\cite{agentdojo}, following the OWASP agentic taxonomy~\cite{owasp_agentic}; every tool identity, permission and response is logged so the attack surface tested is the one deployed~\cite{mcp}.

\paragraph{Oversight.} Because $H\in\sigma$, oversight performance is measured: the \emph{reviewer catch rate} $\rho_{\mathrm{catch}}$ (seeded defects detected / seeded), the \emph{false acceptance rate} (defective outputs accepted / defective), the \emph{unplanned intervention rate} (runs needing intervention outside checkpoints / runs) and review burden (reviewer minutes per completed task).

\begin{table}[t]
\centering\scriptsize
\caption{Metric catalogue. Grading: H human, J calibrated judge, A automatic. Level: lowest autonomy level at which the metric is gated.}
\label{tab:metrics}
\setlength{\tabcolsep}{3pt}
\begin{tabular}{@{}llccc@{}}
\toprule
Family & Metric & Dir. & Grading & Level \\
\midrule
Fidelity & Hallucination rate (Eq.~\ref{eq:hr}) & $\downarrow$ & H,J & A1 \\
 & Citation precision (lenient/strict) & $\uparrow$ & H,J & A1 \\
 & Citation recall & $\uparrow$ & J & A1 \\
Utility & Key-element capture & $\uparrow$ & H,J & A1 \\
 & Irrelevance rate & $\downarrow$ & H,J & A1 \\
 & Completeness (scope attempted) & $\uparrow$ & A & A2 \\
Efficiency & Net time reduction $\Delta_{\mathrm{eff}}$ & $\uparrow$ & A/obs. & A1 \\
 & Cost per successful task & $\downarrow$ & A & A1 \\
Reliability & pass\textasciicircum{}$k$ (Eq.~\ref{eq:passk}); dispersion & $\uparrow$/$\downarrow$ & A,H & A2 \\
 & Recovery rate after tool failure & $\uparrow$ & A & A3 \\
Assurance & Guidance precision / recall & $\uparrow$ & H & A1 \\
 & Policy-violation rate & $\downarrow$ & A,H & A3 \\
 & Escalation appropriateness & $\uparrow$ & H & A3 \\
 & Adversarial incident count & $\downarrow$ & A,H & A3 \\
Oversight & Reviewer catch rate $\rho_{\mathrm{catch}}$ & $\uparrow$ & H (study) & A1 \\
 & False acceptance rate & $\downarrow$ & H & A1 \\
 & Unplanned intervention rate & $\downarrow$ & A & A3 \\
 & Review burden (min/task) & $\downarrow$ & A & A1 \\
\bottomrule
\end{tabular}
\end{table}

%%%%%%%%%%%%%%%%%%%%%%%%%%%%%%%%%%%%%%%%%%%%%%%%%%%%%%%%%%%%%%%%%%%%%%%%%%%%%%%%
\section{Grading and Statistical Protocol}
\label{sec:grading}
\evid{Proposed protocol composed of established methods~\cite{gdpval,mtbench,cohen1960,angelopoulos2023ppi,brown2001interval}}

\paragraph{Blinded expert grading.} Graders are practitioners from the workflow, see model and human artefacts in randomised order without provenance, and score each unit against $\Rb$; each task has $\ge2$ graders and a third adjudicates. Agreement is Cohen's $\kappa=(p_o-p_e)/(1-p_e)$~\cite{cohen1960} or Krippendorff's $\alpha$~\cite{krippendorff2004}; a metric enters a gate decision only if $\kappa\ge0.6$, otherwise the rubric, not the system, is revised. A system's self-assessment is never a gated metric: it violates blinding and inherits judge self-preference~\cite{gu2024judge}.

\paragraph{Calibrated judge with prediction-powered correction.} Expert time is the binding constraint. A judge $f$ scores all $N$ outputs; humans $h$ score a random subset of $n\ll N$; the reported estimate is the prediction-powered estimator~\cite{angelopoulos2023ppi}
\begin{equation}
\hat\theta^{\mathrm{PPI}}=\tfrac1N\textstyle\sum_{i=1}^{N}f(y_i)\;-\;\tfrac1n\textstyle\sum_{i=1}^{n}\big(f(y_i)-h(y_i)\big),
\label{eq:ppi}
\end{equation}
unbiased for the human-graded quantity whatever the judge's accuracy, with variance shrinking as the judge improves. A use case can thus evaluate hundreds of outputs while paying for tens of expert gradings without laundering judge bias into the gate. Judge prompts, model versions and calibration sets are versioned in $\Rb$; trajectory scoring follows Agent-as-a-Judge~\cite{zhuge2024agentjudge} and is confined to what logs can support~\cite{loganalysis2026}.

\paragraph{Sample size and intervals.} To resolve the gap $\delta$ between inner and outer thresholds for a proportion near $\hat p$,
\begin{equation}
n\ \ge\ z_{1-\alpha/2}^{2}\,\hat p(1-\hat p)/\delta^{2};
\label{eq:n}
\end{equation}
for the credit-memo gap of 5 points and $\hat p\approx0.85$, $n\approx196$ units per metric. Units cluster within documents, so cluster-robust or bootstrap intervals over documents are reported; proportions near 0 or 1 use Wilson or Agresti--Coull intervals~\cite{brown2001interval}. Gates are taken on the conservative bound. Tasks are run $K\ge4$ times at deployment settings for A2+ ($\ge2$ for A1); paired comparisons use McNemar or Wilcoxon with Holm correction.

\paragraph{Freeze, parity and stress sets.} $\sigma$ is recorded and held fixed; tuning uses a development split disjoint from the graded set. When two models are compared, everything except $M$ is identical. A small \emph{stress set} probes missing or conflicting sources, ``not determinable'' questions, tool errors and, for A3+, injected instructions; its results are reported separately and gated for assurance only. Algorithm~\ref{alg:eval} states the procedure.

\begin{algorithm}[t]
\caption{\gls{eVal} evaluation of a use case}
\label{alg:eval}
\begin{algorithmic}[1]
\REQUIRE $\U=\langle\T,\Sc,\G,\Rb,a,c\rangle$; candidates $\Sigma=\{\sigma_k\}$; charter $(\thout,\thin,\mathbf w,\nu_{\max})$; $E=E(a,c)$
\ENSURE per-system gate, $\mathrm{EI}$, value estimate, evidence file
\STATE Freeze $\T,\G,\Rb$ and each $\sigma_k$; record versions
\STATE $n\leftarrow$ Eq.~\ref{eq:n}; draw $\T$ stratified; build stress set $\T_s$
\FOR{each $\sigma_k$}
  \FOR{$t_i\in\T\cup\T_s$, $r=1..K(E)$}
    \STATE $(y_{ir},\tau_{ir})\leftarrow\sigma_k(x_i,S_i)$; log $\tau_{ir}$, cost, latency
  \ENDFOR
  \STATE Judge $f$ scores all units; $\ge2$ blinded humans grade a random subset of size $n_h$
  \STATE Compute $\kappa$; \textbf{if} $\kappa_j<0.6$ \textbf{then} revise $\Rb_j$, regrade
  \STATE $\forall j$: $m_j\leftarrow$ PPI (Eq.~\ref{eq:ppi}) or human mean if $n_h\ge n$; $\mathrm{CI}_j\leftarrow$ Wilson / cluster bootstrap
  \STATE pass\textasciicircum{}$k$ (Eq.~\ref{eq:passk}), dispersion; assurance metrics on $\T_s$
  \STATE $(\mathrm{Gate}_k,\mathrm{EI}_k,b_k)\leftarrow$ Algorithm~\ref{alg:gate}
  \STATE $V_k\leftarrow$ Eq.~\ref{eq:value} with $N,c_h,\ell_{\mathrm{err}}$ from charter, $\rho_{\mathrm{catch}}$ from oversight study
\ENDFOR
\STATE Emit scorecards; rank passing systems by EI with sensitivity band; write evidence file; register re-evaluation triggers
\end{algorithmic}
\end{algorithm}

%%%%%%%%%%%%%%%%%%%%%%%%%%%%%%%%%%%%%%%%%%%%%%%%%%%%%%%%%%%%%%%%%%%%%%%%%%%%%%%%
\section{Decision Layer}
\label{sec:decision}
\evid{Proposed}

\begin{figure}[t]
\centering
\includegraphics[width=1.05\columnwidth]{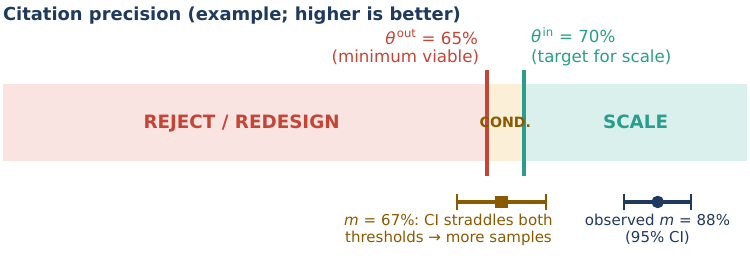}
\caption{\textbf{Two-tier gating.} Decisions use the conservative CI bound; an interval straddling both thresholds calls for more samples, not a verdict.}
\label{fig:thresholds}
\end{figure}

\paragraph{Two-tier gating.} Each gated metric carries two thresholds agreed \emph{before} results are unblinded by the business owner, model owner and second-line risk function. The \emph{outer} threshold $\thout$ is the minimum viable level, anchored to the harm of failure and to current compensating controls; the \emph{inner} threshold $\thin$ is the target at which the use case may scale without additional controls. Between them lies a \emph{conditional} zone: deploy with enhanced review, sampling or scope restriction and a dated commitment to reach $\thin$ (Figure~\ref{fig:thresholds}). Metrics are normalised relative to their thresholds,
\begin{equation}
\nu_j=\mathrm{clip}\!\left(\frac{\sigma_j(m_j-\thout_j)}{\sigma_j(\thin_j-\thout_j)},\,-1,\,\nu_{\max}\right),
\label{eq:nu}
\end{equation}
with $\sigma_j=\pm1$ for higher/lower-is-better, so $\nu_j=0$ at $\thout$, $1$ at $\thin$, negative on failure and capped at $\nu_{\max}$ (3 in the pilot) so one exceptional metric cannot mask a weak one. With $\mathrm{CI}^-_j$ the conservative bound,
\begin{equation}
\mathrm{Gate}(\U)=\begin{cases}
\Scale & \forall j:\ \mathrm{CI}^-_j \succeq \thin_j\\
\Cond & \forall j:\ \mathrm{CI}^-_j \succeq \thout_j,\ \exists j:\ \mathrm{CI}^-_j\prec\thin_j\\
\Reject & \text{otherwise.}
\end{cases}
\label{eq:gate}
\end{equation}
The rule is conjunctive by design: compensation across metrics is appropriate for ranking models, not for deciding whether a bank may rely on a system.

\paragraph{Efficacy Index.} For ranking candidates or versions, $\mathrm{EI}=\sum_j w_j\nu_j$ with charter-fixed weights encoding the risk profile, always reported with a weight-sensitivity band and never as a substitute for the gate. A variant penalising weak dimensions, $\mathrm{EI}_G=\exp(\sum_j w_j\ln(1+\nu_j))-1$ for $\nu_j>-1$, equals zero when every metric sits at $\thout$ and is pulled toward the weakest metric; divergence between EI and $\mathrm{EI}_G$ is a cheap diagnostic of an unbalanced profile. Algorithm~\ref{alg:gate} gives the computation.

\begin{algorithm}[t]
\caption{Two-tier gate and Efficacy Index}
\label{alg:gate}
\begin{algorithmic}[1]
\REQUIRE $m_j,\ \mathrm{CI}^-_j,\ \sigma_j,\ \thout_j,\ \thin_j,\ w_j,\ \nu_{\max}$
\ENSURE Gate, EI, $\mathrm{EI}_G$, binding metric $b$, flags
\STATE Gate $\leftarrow$ \Scale
\FOR{$j=1..J$}
  \STATE $\nu_j\leftarrow$ Eq.~\ref{eq:nu}
  \IF{$\mathrm{CI}^-_j$ fails $\thout_j$} \STATE Gate $\leftarrow$ \Reject
  \ELSIF{$\mathrm{CI}^-_j$ fails $\thin_j$ \textbf{and} Gate $\ne$ \Reject} \STATE Gate $\leftarrow$ \Cond
  \ENDIF
  \IF{$\mathrm{CI}_j$ straddles both thresholds} \STATE flag $j$ under-sampled (Eq.~\ref{eq:n}) \ENDIF
\ENDFOR
\STATE $b\leftarrow\arg\min_j\nu_j$;\ \ $\mathrm{EI}\leftarrow\sum_j w_j\nu_j$;\ \ $\mathrm{EI}_G\leftarrow\exp(\sum_j w_j\ln(1+\nu_j))-1$
\STATE Recompute EI under each alternative $\mathbf w'$ in the charter $\Rightarrow$ sensitivity band
\STATE \textbf{return} Gate, EI, $\mathrm{EI}_G$, $b$, flags
\end{algorithmic}
\end{algorithm}

\paragraph{Value-and-risk model.} For a task performed $N$ times a year at loaded hourly cost $c_h$, with inference cost $c_{\mathrm{infer}}$ per task and expected loss $\ell_{\mathrm{err}}$ if a residual error reaches a decision,
\begin{align}
V_{\mathrm{annual}}&=N c_h\Delta\bar t+V_{\mathrm{quality}}+V_{\mathrm{avoided}}-N c_{\mathrm{infer}}-C_{\mathrm{integr}}\nonumber\\
&\quad-N c_h\bar t_{\mathrm{review}}-C_{\mathrm{rework}}-N\pi_{\mathrm{err}}\ell_{\mathrm{err}},
\label{eq:value}\\
\pi_{\mathrm{err}}&\approx\mathrm{HR}\cdot(1-\rho_{\mathrm{catch}}),
\label{eq:pierr}
\end{align}
where $\Delta\bar t=\bar t_{\mathrm{manual}}-\bar t_{\mathrm{review}}$, $V_{\mathrm{quality}}$ and $V_{\mathrm{avoided}}$ capture outcome improvements and losses avoided, and $C_{\mathrm{integr}}$, $C_{\mathrm{rework}}$ the amortised integration and redo costs; each term is entered as a range and $V_{\mathrm{annual}}$ reported as an interval. Review appears as a cost in Eq.~\ref{eq:value} and as the control on expected loss through $\rho_{\mathrm{catch}}$ in Eq.~\ref{eq:pierr}, so a proposal to ``save time'' by shortening review is automatically charged with the risk it creates. Estimating $\rho_{\mathrm{catch}}$, how often a reviewer reading a fluent draft detects an unsupported number, is the least-studied quantity in the literature; the jagged-frontier result~\cite{dellacqua2023} suggests it may be well below one and fall as trust rises (\S\ref{sec:limits}).

%%%%%%%%%%%%%%%%%%%%%%%%%%%%%%%%%%%%%%%%%%%%%%%%%%%%%%%%%%%%%%%%%%%%%%%%%%%%%%%%
\section{Pilot in a Global Bank}
\label{sec:pilot}
\evid{Documented; sample sizes, CIs and agreement were not recorded in the pilot and are flagged}

The framework was piloted by the innovation function of a global systemically important bank across three workflows spanning finance controls, wholesale credit and operations knowledge management (Table~\ref{tab:portfolio}). Institution, internal platform names and identifiers are anonymised; public model names are retained. Appendix~A (supplementary) gives workflow, prompt and rubric detail.

\begin{table}[t]
\centering\scriptsize
\caption{Pilot portfolio.}
\label{tab:portfolio}
\setlength{\tabcolsep}{3pt}
\begin{tabular}{@{}p{1.6cm}p{1.9cm}p{1.9cm}p{1.9cm}@{}}
\toprule
 & \textbf{A: Control assessment} & \textbf{B: Credit-memo drafting} & \textbf{C: Procedure transformation} \\
\midrule
Level / tier / $E$ & A2 / C3 / E3 & A1 / C3 / E2 & A1 batch / C2 / E2 \\
Inputs & Process--risk--control matrices; control standards & Filings, broker research, internal spreads & Existing procedures (30k corpus, 40k users) \\
Output & Per-control compliance class + recommendation & One of 13 memo sections, cited & Standardised rewrite \\
Models & Gemini~2.5 Flash / Pro & Gemini~2.5, Llama~4 & Pipeline v1, v2 \\
Gated metrics & KEC, IR, HR, guidance P/R, pass\textasciicircum{}4 & CP, KEC, IR, HR, guidance P/R & $\Delta_{\mathrm{eff}}$ \\
Status & Charter set; grading in progress & Human-graded results & Estimated-baseline results \\
\bottomrule
\end{tabular}
\end{table}

\subsection{Use Case A: control design effectiveness (A2)}
Issue quality assurance reviews remediation of high-severity control issues; the assessed step classifies each control in a process--risk--control matrix as compliant, partially compliant or non-compliant against a control standard and rule library. The A2 workflow parses control attributes (who, what, when, where, how), retrieves the standard, emits a per-control judgement table and a structured export. Thresholds and the gold set were fixed before unblinding; results are pending and none are claimed. The charter stage's most useful output was discovering that the original design asked the model to score its own hallucination and coherence; under \gls{eVal} self-assessment is informational only.

\subsection{Use Case B: credit-memo drafting (A1)}
\begin{figure}[t]
\centering
\includegraphics[width=1.04\columnwidth]{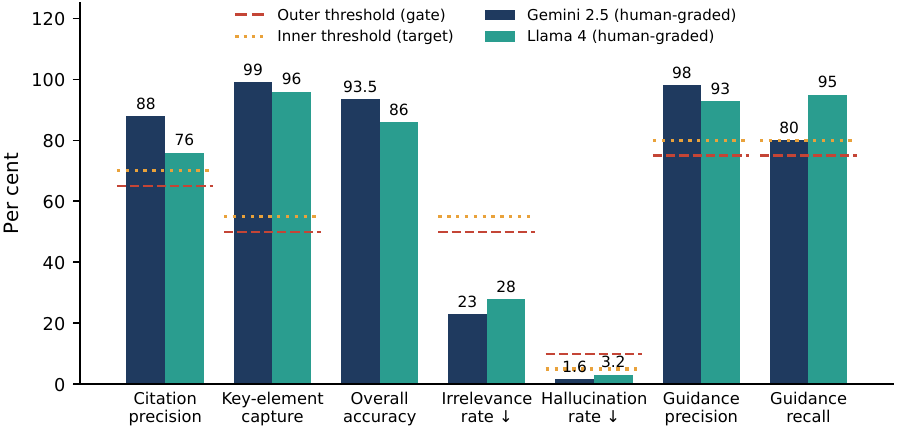}
\caption{\textbf{Use Case B against two-tier thresholds.} Dashed/dotted lines mark outer/inner thresholds.}
\label{fig:ucb}
\end{figure}

Analysts upload filings, research and spreads; the platform drafts one of 13 memo sections with inline citations; the analyst edits; a guidance-verification step checks the draft against credit-policy items. Table~\ref{tab:ucb} and Figure~\ref{fig:ucb} report human-graded results against pre-agreed thresholds. Both models clear every outer threshold. Gemini~2.5 has higher fidelity (88\% CP, 1.6\% HR vs.\ 76\%, 3.2\%), capture and guidance precision; Llama~4 has higher guidance recall (95\% vs.\ 80\%).

\begin{table}[t]
\centering\scriptsize
\caption{Use Case B human-graded results. $\downarrow$ lower is better. Gate column applies Eq.~\ref{eq:gate} to point estimates for illustration; CIs to be added.}
\label{tab:ucb}
\setlength{\tabcolsep}{3pt}
\resizebox{\columnwidth}{!}{%
\begin{tabular}{@{}lccccl@{}}
\toprule
Metric & $\thout$ & $\thin$ & Gemini 2.5 & Llama 4 & Gate \\
\midrule
Citation precision (lenient) & $>$65 & $>$70 & 88.0 & 76.0 & both \Scale \\
Key-element capture & $>$50 & $>$55 & 99.0 & 96.0 & both \Scale \\
Overall accuracy (composite) & -- & -- & 93.5 & 86.0 & not gated \\
Irrelevance rate $\downarrow$ & $<$50 & $<$55$^\dagger$ & 23.0 & 28.0 & both \Scale \\
Hallucination rate $\downarrow$ & $<$10 & $<$5 & 1.6 & 3.2 & both \Scale \\
Guidance precision & $>$75 & $>$80 & 98.0 & 93.0 & both \Scale \\
Guidance recall & $>$75 & $>$80 & 80.0 & 95.0 & G \Cond{}; L \Scale \\
\bottomrule
\end{tabular}}
\par\vspace{2pt}{\scriptsize $^\dagger$Recorded thresholds are inverted (inner should be stricter); retained for fidelity to the pilot and flagged for the next charter.}
\end{table}

\begin{figure}[t]
\centering
\includegraphics[width=1.03\columnwidth]{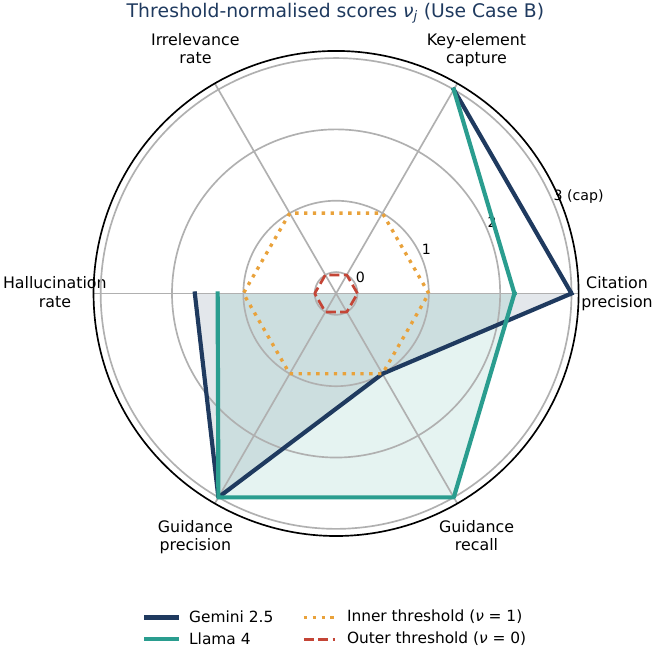}
\caption{\textbf{Threshold-normalised scores $\nu_j$ (Use Case B).} Origin $=\thout$, unit circle $=\thin$, cap 3. Equal-weight and fidelity-weighted EI rank the systems differently.}
\label{fig:radar}
\end{figure}

\paragraph{Interpretation.} Applied mechanically to point estimates, Eq.~\ref{eq:gate} places Llama~4 in \Scale{} on all metrics and Gemini~2.5 in \Cond{} because its recall of 80\% does not strictly exceed 80\%. This is exactly what the framework should expose rather than resolve by fiat: without an interval the distinction is meaningless and the correct action is to grade more guidance items (Figure~\ref{fig:thresholds}). The normalised view (Figure~\ref{fig:radar}) makes a second point: with equal weights and $\nu_{\max}=3$, EI is 2.45 for Gemini and 2.59 for Llama; with fidelity-weighted weights (0.30 HR, 0.25 CP, 0.15 each remaining) the order reverses to 2.62 vs.\ 2.37. Which system to prefer depends on whether the credit function values fewer unsupported statements or fewer missed guidance items, a judgement the charter fixes in advance and the index makes explicit rather than hiding in an average. The pilot did not document harness parity between the two models, so differences cannot be attributed to the models alone, and no timed efficiency comparison was run, so no efficiency result is claimed here.

\subsection{Use Case C: procedure transformation (A1, batch)}
\begin{figure}[t]
\centering
\includegraphics[width=\columnwidth]{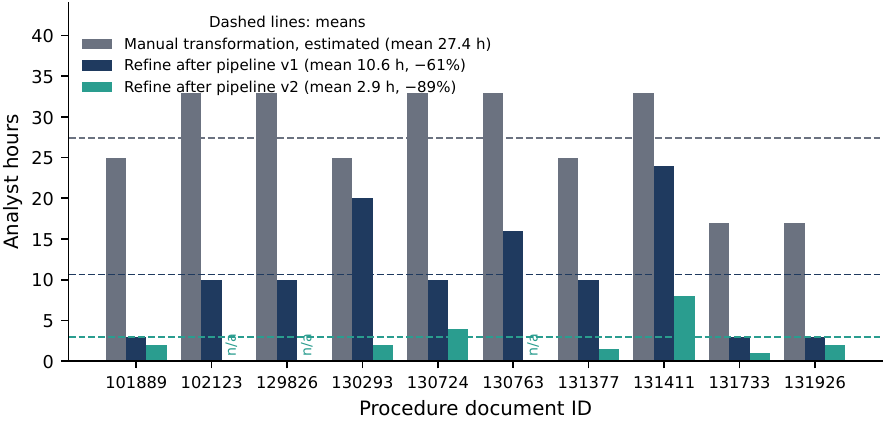}
\caption{\textbf{Use Case C estimated effort by document and pipeline version.}}
\label{fig:ucc}
\end{figure}

Ten frequently used procedures were rewritten by two pipeline versions; owners estimated the effort to refine each output to publishable standard and the effort to transform it manually (Figure~\ref{fig:ucc}). Mean estimated manual effort was 27.4~h/document; refinement after v1 was 10.6~h ($-61\%$) and after v2 2.9~h over the seven documents with v2 outputs ($-89\%$ against the ten-document manual mean; $-88\%$ like-for-like, since the v2 subset's manual mean is 25.0~h). The v1$\to$v2 improvement is consistent within documents, the more credible signal. Three caveats: the baseline is estimated and clusters at three values (17, 25, 33~h), indicating coarse banding subject to recall bias; the v2 mean is on a subset; and fidelity was not graded, so a faster procedure that silently drops a regulatory step would carry negative value under Eq.~\ref{eq:value}.

\subsection{Cross-case lessons and required additions}
Charters force clarity: writing thresholds first exposed an inverted pair, a self-grading step and an unstated baseline. Fidelity and utility trade off across models, and the trade-off is a business judgement to be fixed in weights before results. Efficiency claims are only as good as the baseline. To bring the evidence to the framework's own E2/E3 standard we will: report $n$, cluster-robust 95\% CIs and $\kappa$ for every metric and re-issue gates on conservative bounds; run $K\ge4$ repeats and report pass\textasciicircum{}$\{1,2,4\}$; complete Use Case~A; replace estimated with observed efficiency in a within-subject timed design~\cite{noy2023,dellacqua2023} with four baselines (unaided, assisted, model-only, agentic with matched tooling); add KEC grading to Use Case~C; calibrate a judge and report PPI estimates; re-run B under documented parity with a stress set; and populate Eq.~\ref{eq:value} with $\rho_{\mathrm{catch}}$ elicited by seeding defects into review samples.

%%%%%%%%%%%%%%%%%%%%%%%%%%%%%%%%%%%%%%%%%%%%%%%%%%%%%%%%%%%%%%%%%%%%%%%%%%%%%%%%
\section{Discussion}
\label{sec:discussion}

\paragraph{What the pilot shows and does not show.} \evid{Documented} The framework is \emph{operable}: three business functions in a large regulated institution defined tasks, gold artefacts, metrics and thresholds and, in two cases, produced results a governance forum could read. Frontier models wrapped in retrieval and citation scaffolding reached 1.6--3.2\% hallucination and 96--99\% key-element capture on genuine wholesale-credit documents under human grading, consistent with GDPval~\cite{gdpval}; and standardisation effort fell by an order of magnitude. It does \emph{not} yet show statistically qualified gates, repeated-run reliability, observed efficiency or realised value.

\paragraph{Extending to agentic systems.} \evid{Proposed} As workflows are promoted to A3, a credit agent that gathers filings, computes ratios, drafts sections and reconciles them, four things change. The \emph{unit of grading} becomes the trajectory as well as the artefact: plan validity, retrieval and tool-call correctness, memory reads and writes, checkpoints and final state are logged and graded (Appendix~B (supplementary)), because an agent that reaches a correct memo by an incorrect path is a production liability~\cite{loganalysis2026}. \emph{Reliability becomes gated}, and the vendor-wide plateau~\cite{rabanser2026reliability} implies the outer threshold on pass\textasciicircum{}4 will often bind. \emph{Cost is gated} per successful task~\cite{hal,kapoor2024agents}, since reasoning modes can multiply token cost by an order of magnitude for modest gains~\cite{roig2025kami}. \emph{Assurance expands} to policy violations, escalation and injection incidents~\cite{nistgenai,agentdojo}. Multi-agent pipelines~\cite{guo2024multiagent} additionally require each agent to be evaluated as its own use case, with end-to-end pass\textasciicircum{}$k$ compared against the product of component values to detect interaction effects.

\paragraph{Threats to validity.} \evid{Proposed analysis} \emph{Construct}: sentence-level HR and lenient CP bias in known directions; key-element lists encode one expert view. \emph{Internal}: graders were not always blind; no agreement statistics; harness parity undocumented; estimated baselines. \emph{External}: one institution, one model generation, small task sets - results are about the workflows, not the models, which is why a model change is a re-evaluation trigger. \emph{Statistical}: without $n$ and CIs, Table~\ref{tab:ucb}'s gates illustrate the rule rather than establish findings.

%%%%%%%%%%%%%%%%%%%%%%%%%%%%%%%%%%%%%%%%%%%%%%%%%%%%%%%%%%%%%%%%%%%%%%%%%%%%%%%%
\section{Limitations and Hypotheses}
\label{sec:limits}

The framework presumes a gold artefact can be authored, which holds for generation and classification workflows and less so for open-ended advisory tasks. Thresholds and weights are firm-specific by design, so EI is not comparable across firms; sector reference thresholds are a desirable outcome the framework enables but cannot supply. The conjunctive gate is conservative and its rejection rate grows with the number of gated metrics, a multiplicity effect the charter manages by gating only metrics mapped to a material harm. The agentic extension is specified but not yet exercised.

\evid{Hypothesis} We state testable conjectures. \textbf{H1} $\rho_{\mathrm{catch}}$ for fluent, cited drafts is below 0.7 and declines as measured HR falls (seeded-defect blinded review study). \textbf{H2} Thresholds agreed independently by peer institutions fall in a narrow band, enabling sector references (multi-institution charter on a synthetic task set). \textbf{H3} A calibrated judge reaches $\kappa\ge0.6$ on HR and IR but not on KEC or value-add. \textbf{H4} For A3 agents on document workflows, pass\textasciicircum{}4 at $\thout$ binds more often than any fidelity metric. \textbf{H5} Use cases passing the inner gate show measurable capacity or P\&L effects within two quarters at a rate well above the 5\% reported for pilots generally~\cite{mit_divide} (prospective registration; the predictive-validity standard of~\cite{predictive_validity}). \textbf{H6} Conjunctive gating produces fewer approvals later reversed than any weighted average of the same metrics, at the cost of more \Cond{} outcomes (retrospective portfolio classification). \textbf{H7} pass\textasciicircum{}$k$ declines with human-equivalent task duration~\cite{metr2025horizon} unless checkpoint density scales with it. Ablations over the components of $\sigma$ (retrieval, scaffold, tools, model, autonomy, review design) are a further priority, given KAMI's finding that tool-message changes alone move accuracy and cost by large factors~\cite{roig2025kami}. We plan an open reference implementation of Algorithms~\ref{alg:eval}--\ref{alg:gate} with a schema for Eqs.~\ref{eq:usecase}--\ref{eq:config} and a synthetic public task set.

%%%%%%%%%%%%%%%%%%%%%%%%%%%%%%%%%%%%%%%%%%%%%%%%%%%%%%%%%%%%%%%%%%%%%%%%%%%%%%%%
\section{Conclusion}
Enterprise GenAI has an evaluation problem before it has a capability problem. Public benchmarks show what frontier models can do; they cannot tell a bank whether a specific workflow is fit, reliable, safe and worth scaling on its own documents and under its own controls, and the internal evaluations that try rarely meet the standard a risk committee or a reviewer requires. \gls{eVal} is an evaluation system for closing that gap: a formal, frozen evaluation object; a metric catalogue spanning fidelity, utility, efficiency, reliability, assurance and oversight; a grading protocol that scales expert judgement with calibrated automation without laundering its bias; a conjunctive two-tier gate on conservative bounds; and a value model that puts review effort and residual risk on the same ledger as time saved. A pilot in a global bank shows the system to be operable and yields encouraging fidelity and efficiency results, while making plain the evidence still required. We offer the framework, pilot and hypotheses in the expectation that enterprise AI evaluation will mature as model benchmarking has - through shared definitions, pre-registered thresholds and honest reporting of uncertainty.

%%%%%%%%%%%%%%%%%%%%%%%%%%%%%%%%%%%%%%%%%%%%%%%%%%%%%%%%%%%%%%%%%%%%%%%%%%%%%%%%
\bibliography{main}
\bibliographystyle{mlsys2025}

\appendix 

\appendix

% \section*{Supplementary Appendix}
\label{app:appendix}

\section{Pilot Detail}
\label{app:pilot}

\subsection*{A. Evaluation lifecycle}
\begin{figure}[H]
\centering
\includegraphics[width=1.04\columnwidth]{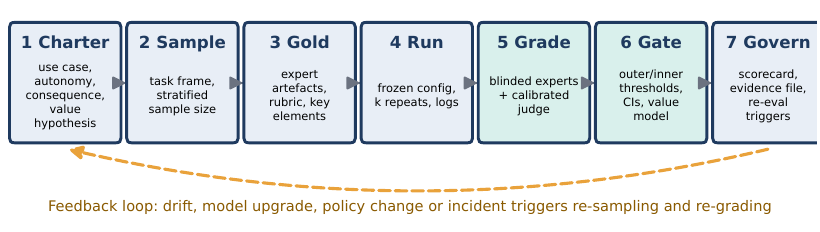}
\caption{Evaluation lifecycle with re-evaluation feedback loop.}
\label{fig:lifecycle}
\end{figure}

Figure~\ref{fig:lifecycle} shows the seven-stage lifecycle that Algorithm~1 of the main paper formalises; stages 5 (grade) and 6 (gate) are where internal evaluations are typically weakest.

\subsection{Autonomy ladder}
\begin{figure}[H]
\centering
\includegraphics[width=1.04\columnwidth]{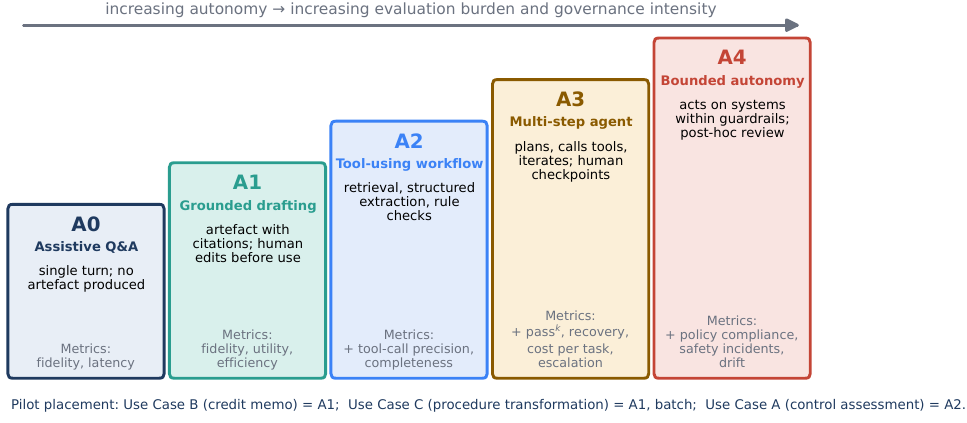}
\caption{Autonomy levels A0--A4 and the metric families they require.}
\label{fig:ladder}
\end{figure}

Figure~\ref{fig:ladder} expands the five autonomy levels of \S3.4 and the metric families each adds; Table~\ref{tab:agentic} lists what changes qualitatively when a workflow becomes agentic.

\begin{table}[H]
\centering\scriptsize
\caption{What changes when a workflow becomes agentic, and what must therefore be measured.}
\label{tab:agentic}
\setlength{\tabcolsep}{3pt}
\begin{tabular}{@{}p{1.3cm}p{1.6cm}p{1.8cm}p{2.6cm}@{}}
\toprule
Dimension & Assistive (A0--A1) & Agentic (A3--A4) & Evaluation implication \\
\midrule
Task control & User-led & System plans and decomposes & Plan validity, stopping rule, recovery \\
Context & Prompted / uploaded & Dynamic retrieval and tool calls & Evidence coverage, retrieval and tool-call correctness \\
Action & Text only & State-changing actions & Authorisation, reversibility, action-safety gates \\
State & Short-lived & Memory across steps & State consistency; memory poisoning tests \\
Failure mode & Wrong content & Wrong content and wrong action & Measure separately \\
Human role & Reviewer / editor & Approver, supervisor & Review load, intervention rate, escalation quality \\
Economic unit & Tokens, latency & End-to-end workflow cost & Cost per \emph{successful} task incl.\ human effort \\
Attack surface & Prompt content & Prompt, tools, memory, inter-agent messages & Injection, tool misuse, privilege abuse, exfiltration \\
\bottomrule
\end{tabular}
\end{table}

\subsection{Use Case A: workflow phases}
\begin{table}[H]
\centering\scriptsize
\caption{Use Case A: phases, AI contribution and applicable metrics.}
\setlength{\tabcolsep}{3pt}
\begin{tabular}{@{}p{1.6cm}p{2.7cm}p{2.9cm}@{}}
\toprule
Phase & AI contribution & Metrics \\
\midrule
Identification & Extract issue facts, stakeholders, corrective-action context & Extraction KEC; omission rate; citation precision \\
Evaluation / scoping & Assess reasonableness of remediation plans; completeness of controls & KEC; HR; recommendation validity \\
Remediation design & Classify control design vs.\ standard and rule library; recommend fixes & Classification P/R vs.\ frozen gold; pass\textasciicircum{}$k$; completeness \\
Sustainability & Assess whether controls are implemented and operating & Evidence sufficiency; escalation appropriateness \\
\bottomrule
\end{tabular}
\end{table}

The system specification includes two guardrails that are part of the evaluation object: no analysis begins before the input file is present and the analyst explicitly instructs the run, and outputs are exported to a structured deliverable rather than free text. Thresholds for KEC, IR, HR, guidance precision/recall and classification pass\textasciicircum{}4 were left as ``to be determined'' at charter until the control owner fixed them before unblinding; grading of Gemini~2.5 Flash and Pro outputs was in progress at submission.

\subsection{Use Case B: generalised drafting instruction}
The following is the drafting instruction for the \emph{source of repayment} memo section, generalised for publication. It illustrates why prompts are part of the evaluation object: the prohibition in item~4 is a policy rule whose violation is an assurance failure, and the accounting-period requirement in item~1 defines key elements the gold specification must enumerate.

\begin{quote}\small
\emph{Role.} You are a corporate banking analyst preparing the source-of-repayment section of a credit memo for a corporate borrower seeking a revolving credit facility.

\emph{Objective.} Identify and document the primary, secondary and other sources of repayment with specific numerical values and explicit accounting periods.
\begin{enumerate}\setlength\itemsep{0pt}
\item State the accounting periods used (e.g., LTM ended [date], FY[YYYY], three-year average).
\item Align repayment sources with the debt structure, distinguishing short- and long-term tenors.
\item Use consistent currency units and cite source materials explicitly (financial statements, base-case model, appraisals).
\item \textbf{Never} present revolving credit facilities, undrawn lines or short-term borrowings as a source of repayment.
\item \emph{Primary source:} free cash flow (CFO less capex) relative to scheduled amortisation.
\item \emph{Secondary source:} refinancing capacity, collateral realisation (asset values, LTV, advance rates) or pending transactions.
\item \emph{Other sources:} realistic alternatives aligned to debt tenor.
\end{enumerate}
\end{quote}

\subsection{Use Case C: per-document effort}
\begin{table}[H]
\centering\scriptsize
\caption{Estimated effort per procedure (hours). n/a: no v2 output assessed.}
\label{tab:ucc}
\begin{tabular}{@{}llrrr@{}}
\toprule
Doc & Title (abbrev.) & Manual & v1 & v2 \\
\midrule
101889 & Security word rules & 25 & 3 & 2 \\
102123 & Do-not-release to any caller & 33 & 10 & n/a \\
129826 & Inactive and dormant accounts & 33 & 10 & n/a \\
130293 & Verifications and overrides & 25 & 20 & 2 \\
130724 & Travel suppression & 33 & 10 & 4 \\
130763 & Vulnerable adult incident report & 33 & 16 & n/a \\
131377 & Regulation DD & 25 & 10 & 1.5 \\
131411 & Regulation E & 33 & 24 & 8 \\
131733 & Consumer fairness policy & 17 & 3 & 1 \\
131926 & Branch investment procedures & 17 & 3 & 2 \\
\midrule
\multicolumn{2}{@{}l}{Mean ($n{=}10$)} & 27.4 & 10.6 & -- \\
\multicolumn{2}{@{}l}{Mean (v2 subset, $n{=}7$)} & 25.0 & 10.4 & 2.9 \\
\bottomrule
\end{tabular}
\end{table}

\subsection{Grading rubric templates}
\begin{table}[H]
\centering\scriptsize
\caption{Sentence-level fidelity and relevance rubric (Use Cases A, B).}
\setlength{\tabcolsep}{3pt}
\begin{tabular}{@{}p{1.4cm}p{4.3cm}p{1.4cm}@{}}
\toprule
Field & Instruction to grader & Values \\
\midrule
Assertive? & Does the sentence make a checkable factual claim? Headings, questions and explicit recommendations are not assertive. & Y / N \\
Supported? & For each claim, can you locate support in the permitted sources? Support from your own knowledge does \emph{not} count. & Fully / Partially / Not \\
Relevant? & Does the sentence contribute to the instruction for this section? & Y / N \\
Citation used? & Is the cited chunk's content actually used (strict) or at least reflected (lenient)? & Strict / Lenient / Not \\
Key elements & Tick each gold key element present and correct; note incorrect presentations. & checklist \\
Value-add & Correct, relevant content absent from the human baseline. & 0 / 1 / 2 \\
\midrule
\multicolumn{3}{@{}l}{\emph{Guidance-verification rubric}} \\
Flag correct? & For each flagged guidance item, does the draft in fact breach or omit it? & TP / FP \\
Missed item? & For each mandatory item not flagged, does the draft breach or omit it? & FN / correct \\
Severity & Business-assigned severity of the item. & Low / Med / High \\
\bottomrule
\end{tabular}
\end{table}

\subsection*{B. Trajectory-Level Evaluation Record}
\label{app:traj}
\begin{table}[H]
\centering\scriptsize
\caption{Trajectory-level record for A3--A4 systems; each layer is logged per run and graded.}
\setlength{\tabcolsep}{3pt}
\begin{tabular}{@{}p{1.4cm}p{2.2cm}p{1.9cm}p{1.6cm}@{}}
\toprule
Layer & Recorded & Metrics & Failure example \\
\midrule
Input / context & Source identities, completeness, access policy applied & Completeness; policy-violation rate & Required doc missing; restricted source accessed \\
Planning & Sub-tasks, dependencies, stopping rule & Plan validity (judge); step count & Unsafe plan; non-termination \\
Retrieval / memory & Queries, documents, ranks, evidence used; memory reads/writes & Retrieval precision; evidence coverage; poisoning tests & Correct source not retrieved; stale memory \\
Tool use & Tool, arguments, authorisation, result, retries & Tool-call precision; recovery rate; unauthorised-call rate & Wrong entity; action outside permission \\
Synthesis & Claim--evidence linkage & HR; CP; KEC & Unsupported claim carried forward \\
Human oversight & Checkpoint, intervention reason, action, time & Unplanned intervention rate; escalation appropriateness; $\rho_{\mathrm{catch}}$ & Reviewer misses defect; checkpoint bypassed \\
Final outcome & Acceptance criteria; objective state check & Task success; pass\textasciicircum{}$k$; cost per successful task & Mandatory element missing; state not as claimed \\
\bottomrule
\end{tabular}
\end{table}

\subsection*{C. Governance Mapping}
\label{app:gov}
\begin{table}[H]
\centering\scriptsize
\caption{Mapping of EnterpriseVal artefacts to governance requirements.}
\setlength{\tabcolsep}{2.5pt}
\begin{tabular}{@{}p{1.7cm}p{1.4cm}p{1.5cm}p{1.4cm}p{1.2cm}@{}}
\toprule
Artefact & SR 11-7 / SS1/23 & NIST AI RMF + GenAI Profile & EU AI Act (high-risk) & ISO/IEC 42001 \\
\midrule
Charter (Eq.~1), $a$, $c$, thresholds & Model definition, intended use & Map: context, tolerances & Art.~9 risk mgmt & Cl.~6 planning; impact assessment \\
Gold set, sampling frame & Development data documentation & Measure: test-data representativeness & Art.~10 data governance & Annex A data controls \\
Grading records, $\kappa$, judge calibration & Independent validation evidence & Measure: metric validity, human evaluation & Art.~15 accuracy, robustness & Cl.~9 performance evaluation \\
Gate decision, scorecard & Validation outcome, conditions of use & Manage: risk treatment & Art.~14 human oversight & Cl.~9.3 management review \\
Value-and-risk model & Materiality assessment & Govern: benefit--risk & -- & Cl.~6.1 objectives \\
Re-evaluation triggers, monitoring & Ongoing monitoring, change control & Manage: post-deployment monitoring & Art.~72 post-market; Art.~12 logging & Cl.~10 improvement \\
\bottomrule
\end{tabular}
\end{table}

\subsection*{D. Threshold-Setting Protocol}
Thresholds are set in a facilitated session before any results are unblinded, with the business owner, model owner, second-line risk function and evaluation lead present.
\begin{enumerate}\setlength\itemsep{0pt}
\item \textbf{Enumerate harms.} For each candidate metric, describe the concrete harm of failure. Metrics with no material harm are informational and not gated.
\item \textbf{Anchor $\thout$ to harm and compensating controls.} At what level does residual risk, given current review design, become unacceptable?
\item \textbf{Anchor $\thin$ to scale.} At what level would the risk function accept lighter compensating controls? The gap should be at least the expected sampling uncertainty at the planned $n$ (Eq.~7).
\item \textbf{Fix weights and $\nu_{\max}$}, with rationale and at least one alternative weighting for sensitivity reporting.
\item \textbf{Sign and version the charter.} Any post-unblinding change is a recorded deviation.
\end{enumerate}

\subsection*{E. Value Model Decomposition}
\begin{figure}[H]
\centering
\includegraphics[width=1.12\columnwidth]{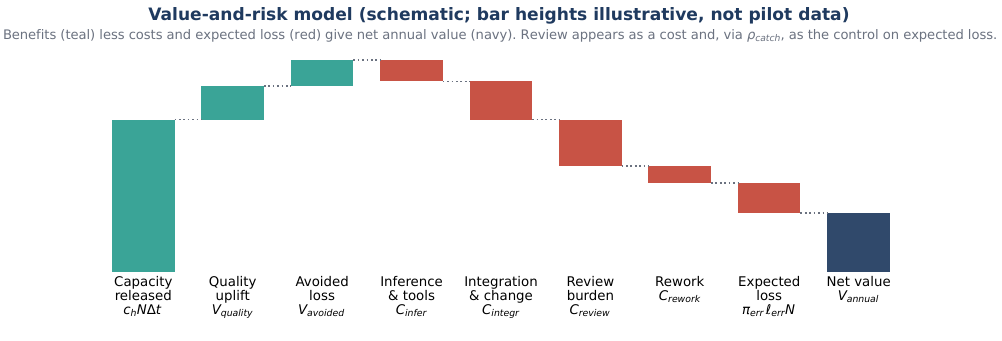}
\caption{Schematic decomposition of Eq.~11 (main paper). Bar heights are illustrative, not pilot data.}
\end{figure}

\subsection*{F. Scorecard Template}
\begin{table}[H]
\centering\scriptsize
\caption{One scorecard per use case per cycle; every field is mandatory.}
\begin{tabular}{@{}p{7.6cm}@{}}
\toprule
Use case ID; business owner; model owner; second-line reviewer \\
Business outcome and value hypothesis \\
Task specification $\mathcal T,\mathcal G,\mathcal R$ (versions) \\
Autonomy level $a$, consequence tier $c$, intensity $E(a,c)$ \\
Configuration $\sigma=\langle M,P,\Phi,U,\Gamma,H\rangle$ with versions \\
Baseline workflow and measurement mode (estimated / observed) \\
Gated metrics: estimate, CI, $\thout$, $\thin$, $\nu_j$, grading mode, $\kappa/\alpha$ \\
Reliability: $K$, pass\textasciicircum{}1/2/4, dispersion \\
Assurance and stress-set results \\
Oversight: $\rho_{\mathrm{catch}}$, FAR, intervention rate, review burden \\
Sample sizes (units, documents, tasks); under-sampled flags \\
Gate decision and binding metric $b$ \\
EI and $\mathrm{EI}_G$ with weight-sensitivity band \\
Value-model inputs ($N$, $c_h$, $\ell_{\mathrm{err}}$, $\rho_{\mathrm{catch}}$) and $V_{\mathrm{annual}}$ interval \\
Observed failure modes; conditions of use; compensating controls \\
Re-evaluation triggers and next review date \\
\bottomrule
\end{tabular}
\end{table}

% \bibliography{main}\bibliographystyle{mlsys2025}
% \end{document}

\end{document}